\documentclass[conference,compsoc]{IEEEtran}

\usepackage{cite}
\usepackage{times}  
\usepackage{helvet} 
\usepackage{courier}  
\usepackage{url}  
\usepackage{graphicx} 
\usepackage{caption} 
\usepackage{algorithm}
\usepackage{algorithmic}

\usepackage{booktabs}
\usepackage{graphicx}
\usepackage{adjustbox}
\usepackage{makecell}

\usepackage{placeins}
\usepackage{stfloats}

\usepackage{hyperref}

\usepackage{newfloat}
\usepackage{listings}
\DeclareCaptionStyle{ruled}{labelfont=normalfont,labelsep=colon,strut=off} 
\floatstyle{ruled}
\newfloat{listing}{tb}{lst}{}
\floatname{listing}{Listing}

\usepackage{xspace} 
\usepackage{xcolor}
\usepackage{subcaption}

\newcommand{\ContractEval}{\textit{ContractEval}\xspace}

\usepackage{longtable}
\usepackage{enumitem}

\title{ContractEval: Benchmarking LLMs for Clause-Level Legal Risk Identification in Commercial Contracts}

\setlist[itemize]{leftmargin=*}

\begin{document}

\author{
\IEEEauthorblockN{
Shuang Liu\IEEEauthorrefmark{1},
Zelong Li\IEEEauthorrefmark{2},
Ruoyun Ma\IEEEauthorrefmark{3},
Haiyan Zhao\IEEEauthorrefmark{4},
Mengnan Du\IEEEauthorrefmark{4},\\
}
\IEEEauthorblockA{\IEEEauthorrefmark{1}Carnegie Mellon University}
\IEEEauthorblockA{\IEEEauthorrefmark{2}Rutgers University}
\IEEEauthorblockA{\IEEEauthorrefmark{3}Stanford University}
\IEEEauthorblockA{\IEEEauthorrefmark{4}New Jersey Institute of Technology}
}

\maketitle

\begin{abstract}
The potential of large language models (LLMs) in specialized domains such as legal risk analysis remains underexplored. In response to growing interest in locally deploying open-source LLMs for legal tasks while preserving data confidentiality, this paper introduces  \ContractEval, the \emph{first} benchmark to thoroughly evaluate whether open-source LLMs could match proprietary LLMs in identifying clause-level legal risks in commercial contracts. Using the Contract Understanding Atticus Dataset (CUAD), we assess 4 proprietary and 15 open-source LLMs. Our results highlight five key findings: (1) Proprietary models outperform open-source models in both correctness and output effectiveness, though some open-source models are competitive in certain specific dimensions. (2) Larger open-source models generally perform better, though the improvement slows down as models get bigger. (3) Reasoning (``thinking”) mode improves output effectiveness but reduces correctness, likely due to over-complicating simpler tasks. (4) Open-source models generate “no related clause” responses more frequently even when relevant clauses are present. This suggests ``laziness" in thinking or low confidence in extracting relevant content. (5) Model quantization speed up inference but at the cost of performance drop, showing the tradeoff between efficiency and accuracy. These findings suggest that while most LLMs perform at a level comparable to junior legal assistants, open-source models require targeted fine-tuning to ensure correctness and effectiveness in high-stakes legal settings. \ContractEval offers a solid benchmark to guide future development of legal-domain LLMs. The code for \ContractEval can be found at \url{https://github.com/olivialiu121/ContractEval}.
\end{abstract}

\section{Introduction}
Managing legal risk is central to successful commercial transactions. For instance, imagine a technology company plans to buy a video game company. Before the deal goes through, the buyer's legal team needs to review the seller’s contracts to check if any existing or potential liability that harms the transaction. They may look into intellectual property (IP) agreements to confirm IP ownership, shareholder agreements to assess potential liabilities, and service contracts to spot any problems that could affect the deal. Such contract review helps buyer understand any legal issues or obligations that come with the company they want to buy. If any critical issuers are spotted, the buyer could lower the purchase price or even terminate the transaction. 

Despite the practical importance, contract review is time-consuming, costly, and even suffering for both lawyers and clients \cite{ldd}. A common first step of contract review is asking junior legal assistants to manually extract clauses related to specific risks or concerns (e.g. extracting clauses discussing who owns a copyright or patent, or highlighting when a contract starts/ends) \cite{CUAD}. This process can cost hundreds of thousands of dollars, sometimes merely to confirm there is no major problem blocking the transaction. Although the task is simple, reviewing contracts is slow and expensive because of the large number of documents and the manual effort involved. The potential of large language models (LLMs) in specialized domains like law remains widely discussed but underexplored. Most prior work has focused on tasks such as legal case retrieval, judgment prediction, and legal text generation \cite{xiao2018cail2018, court, Case}, while contract review and legal risk assessment have been largely overlooked. Notably, contract review is fundamentally different from legal case retrieval and other legal text generation tasks. While case retrieval foucs on document-level retrieval without requiring locating spans in the given context, contract review involves extracting exact sub-strings in the given context. Therefore, contract review puts higher requirements on span-level understanding and domain-specific knowledge.

The lack of a comprehensive assessments of state-of-the-art LLMs prevents integrating of LLMs into contract review workflows. On the other hand, law firms also face strict obligations to protect client confidentiality \cite{ABA}, which makes them cautious of transmitting sensitive data to external servers. This has led to growing interest in locally deploying open-source LLMs to reduce data exposure and meet privacy compliance rules. This raises a pressing question: \emph{Can open-source LLMs automate contract review while maintaining data privacy and matching the performance of proprietary models?} Despite its practical importance, this question remains unanswered due to the lack of detailed testing in the legal area.
 
In this paper, we introduce \ContractEval, the \emph{first} benchmark for systematically evaluating both open-source and proprietary LLMs on contract review tasks. \ContractEval includes 19 leading models and covers 41 legal risk categories commonly assessed during contract review. \ContractEval is built on the test data of Contract Understanding Atticus Dataset (CUAD) \cite{CUAD}. The benchmark measures performance across three key dimensions: correctness (F1 scores), output effectiveness and accuracy (Jaccard similarity), and “laziness” (rate of incorrect “no related clause” responses). Our results highlight both the strengths and current limitations of open-source LLMs  and provide actionable insights for applying LLMs to the legal area: (1) Proprietary models outperform open-source ones in both correctness and output effectiveness. Proprietary models show more balanced performance across both correctness and effectiveness, while open-source models tend to perform well in one but not both. (2) Model size has diminishing returns. While proprietary models' performance is not size-dependent, open-source models benefit from scaling, but the gains slow down beyond the optimal size. This makes size-efficiency trade-offs important for contract review tasks. (3) LLM thinking mode improves output effectiveness and conciseness but hurts correctness, likely due to over-complicating simpler tasks. This trade-off suggests that lawyers should decide on thinking strategies based on whether they prioritize conciseness or accuracy, and pay closer attention to the other one. (4) LLM quantization reduces GPU costs but slightly lowers performance, especially in thinking mode. Unless resource constraints are strict, full-precision models may be preferable for legal tasks. (5) Model performance varies across legal categories, with significantly lower correctness in less common or longer clauses. This highlights the need for domain-specific fine-tuning to address the imbalance among categories.
Our major contributions are as follows:
\begin{itemize}
    \item We propose \ContractEval,  the \emph{first} benchmark to assess 19 state-of-the-art proprietary and open-source LLM on clause-level contract review.
    \item Our benchmarking provides practical insights for the legal industry: open-source models have potential but currently fall behind in correctness, conciseness, and handling category imbalance. Fine-tuning should improve their correctness and information retrieval capability for longer or less common clause categories. 
    \item Our benchmarking bridges the LLM research communities and legal industry. It supports the development of domain-specific LLMs and encourages further future works on remaining questions (e.g. how to design better human-based evaluation metrics and ensure LLM output reliability and consistency). 
\end{itemize}

\section{Related Work}
\label{relatedwork}
\begin{figure*}[t]
    \centering
    \includegraphics[width=\linewidth]{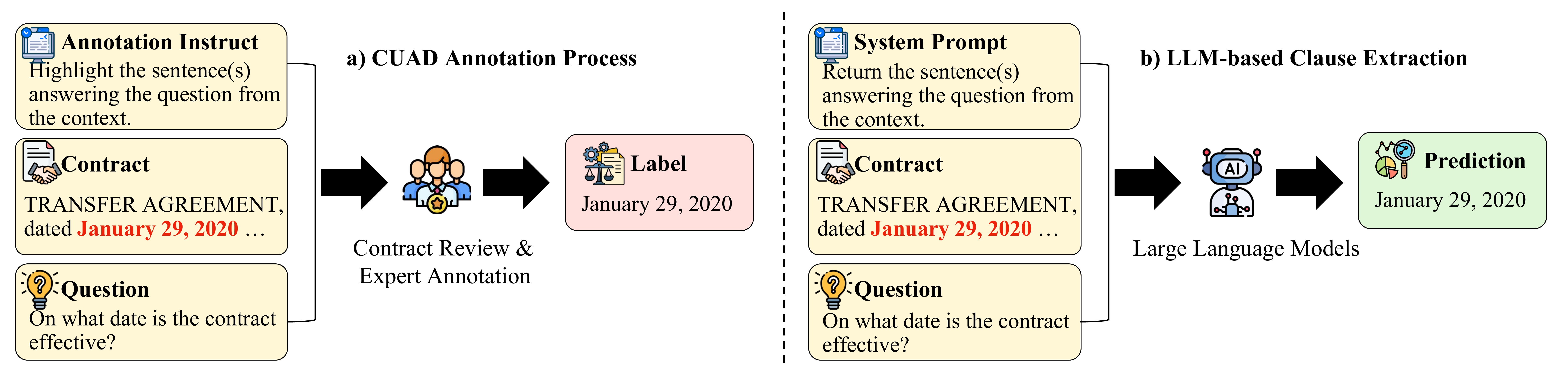}
    \caption{Workflow of Contract Review Implementation}
    \label{fig:workflow}
\end{figure*}


\subsection{AI Legal Assistance and Benchmarks}
With the growing application of NLP in the legal domain, several datasets and benchmarks have been established to evaluate model performance on legal tasks. Most of such works focus on five areas: (1) court judgments, such as VICTOR, which analyzes Brazilian Supreme Court documents \cite{victor}, and LAWMA, which provides legal annotations of U.S. Supreme Court opinions \cite{lawma}; (2) specific sub-types of legal documents, such as wills \cite{will} and standardized German consumer contracts in AGB-DE \cite{braun2024agb}; (3) legal reasoning, such as CHANCERY, which assesses corporate governance reasoning \cite{irwin2025chancery}; (4) legal case retrieval, such as legal RAG \cite{reasoningcaseretrieval}; and (5) general legal knowledge, such as LawBench, which evaluates LLMs on a wide range of legal tasks \cite{fei2023lawbench}. 

Notably, contract review and clause extraction differ significantly from legal case retrieval and document generation. Contract review and clause extraction require span-level predictions, requiring models pinpoint exact clauses or sentences within a given context (e.g. a long contract). However, legal case retrieval focuses on document-level relevance, i.e., finding the most relevant prior decisions or cases, in stead of identifying exact text spans in the given context. On the other hand, judgment generation and legal text generation falls under generation tasks, instead of classification or information retrieval task. As a result, contract review task requires span-level understanding and domain-specific knowledge, making it more difficult and more specialized. However, contract review and legal risk identification still remain underexplored. The most relevant datasets and benchmarks, CUAD \cite{CUAD} and ContractNLI \cite{koreeda2021contractnli}, focus on clause-level annotation tasks but mainly evaluate earlier models such as BERT~\cite{devlin2019bert}. With the fast progressof LLMs, there is a clear need for an up-to-date benchmark that tests both proprietary and open-source models on clause-level legal risks in contract review tasks.

\section{\ContractEval Setup and Implementation}
In this section, we outline the experimental setup and implementation details of \ContractEval for testing LLMs on clause-level legal risk identification in contract review.

\subsection{Dataset Description}
CUAD is an expert-annotated dataset built for contract review task. It provides high-quality labels across 41 clause types from real-world contracts. CUAD is developed to test how well models can find clauses that answer legal questions or related to a certain legal risk. For instance, it's important in real life to clearly identify the "renewal term" of a contract \cite{CUAD}, as the company needs to prepare renewal documents in advance. In this work, we use CUAD's test set, which contains 4,128 data points across 41 clause categories from 102 unique contracts. Each data point corresponds to a (contract, question) pair with a specific clause-type label. For negative cases (i.e. questions with no relevant clause in the contract), the length of the label is 0. Table~\ref{CUAD sample} provides three sample entries, including the label, category, and detailed question used to query the model.

\begin{table}[t]
\centering
\caption{Summary of CUAD Test Data}
\begin{tabular}{@{}ll@{}}
\toprule
\textbf{Metric} & \textbf{Value} \\
\midrule
Total Data Points & 4,128 \\
Context Length Range & 0.6k - 301k characters \\
Label Type & Span in context (clause) \\
Label Distribution & 30\% positive / 70\% negative \\
\bottomrule
\end{tabular}
\label{tab:cuadsummary}
\end{table}


\begin{table*}[t]
\centering
\caption{Example categories with detailed question and sample label annotated by experts}
\label{CUAD sample}
\small 
\begin{tabular}{@{}p{0.15\textwidth}p{0.3\textwidth}p{0.5\textwidth}@{}}
\toprule
\textbf{Category} & \textbf{Detailed Question} & \textbf{Label: Sample Answer} \\
\midrule
Effective Date & On what date is the contract effective? & August 10, 2007 \\
Renewal Term & What is the renewal term after the initial term expires? & Thereafter the term of this Agreement shall automatically renew for consecutive periods of two (2) years each. \\
License Grant & Does the contract contain a license granted by one party to its counterparty? & [] (Note: empty label indicating no relevant clauses) \\
\bottomrule
\end{tabular}
\end{table*}

CUAD is well-suited for our study for two key reasons. First, the dataset features high-quality contracts and annotations. CUAD collects the contracts from the Electronic Data Gathering, Analysis, and Retrieval (EDGAR) system \cite{CUAD}. These contracts are based on real transactions, covering a broad range of clause types. Annotations were conducted by law students with 70–100 hours of specialized training, under the supervision of experienced lawyers. This training process ensures a high level of annotation accuracy and legal relevance. Second, CUAD closely matches the goal of this benchmark: identifying legal risk by extracting clauses from different contracts. As discussed in Section~\ref{relatedwork}, other contract datasets either focus on specific legal domains (e.g., wills  or standardized consumer contract) or are annotated in languages other than English \cite{irwin2025chancery, braun2024agb, will}. ContractNLI is document-level dataset involving classifying a hypothesis are entailed by, contradictory to, or not mentioned in a given contract\cite{koreeda2021contractnli}. However, this task differs significantly from real like contract review. In contrast, CUAD was built specifically for clause-level legal risk identification in real-world contracts, making it better suited for our task.

\subsection{Selection of Models} 
In this paper, we evaluate 19 LLMs, including 4 proprietary models: GPT 4.1 \cite{gpt4.1}, GPT 4.1 mini \cite{gpt4.1mini}, Gemini 2.5 pro preview \cite{comanici2025gemini}, Claude sonnet 4 \cite{claudesonnet4} and 15 open-source models. We categorize the open-source models into four groups: 
\begin{itemize}
    \item DeepSeek series: DeepSeek R1 Distill Qwen 7B and DeepSeek R1 0528 Qwen3 8B \cite{guo2025deepseek}
    \item LLaMA series: LLaMA 3.1 8B Instruct \cite{dubey2024llama}
    \item Gemma series: Gemma 3 4B and 12B \cite{team2025gemma}
    \item Qwen3 series (for each, we test both enabling and disabling thinking mode): Qwen3 4B, Qwen3 8B, and Qwen3 14B, Qwen3 8B AWQ, and Qwen3 8B FP8 \cite{yang2025qwen3}
\end{itemize}
We use three main criteria to guide model selection:
\begin{itemize}
    \item \textbf{Maximum token capability}. Since the input data consists of long and complex real-world legal contracts, models need to handle large contexts.  As shown in Table~\ref{tab:cuadsummary}, the longest contract in the CUAD test set has over 300k characters. Some legal-specific LLMs, such as SaulLM-7B with a 32k token limit \cite{colombo2024saullm}, are still not suitable for our task. To ensure the ability to process such documents effectively, we include only models with a maximum context window of at least 128k tokens which fully covers the longest contract.

    \item \textbf{Latest models}. To evaluate the performance of state-of-the-art LLMs, we select models released around mid-2025. This timeframe captures the most recent advancements in both proprietary and open-source models. For instance, the Qwen3 series by Alibaba was released in April 2025. Focusing on recently released models ensures our evaluation reflects the cutting edge of current LLM capabilities.
    \item \textbf{Practical benefits of open-source models}. We focus on open-source models for two key reasons. First, as noted above, legal service providers are subject to strict obligations to protect client confidentiality and minimize data transmission to external servers. This makes locally deployed, open-source models significantly more attractive for real-world legal assistance tasks. Comparing their performance to proprietary models will provide practically relevant insights into their readiness for deployment in sensitive legal environments. Second, open-source models offer clear advantages in cost efficiency. For example, using GPT-4.1 incurs a charge of \$2 per million input tokens and \$8 per million output tokens \cite{gptprice}. Running GPT-4.1 on the CUAD test set would cost approximately \$50. In contrast, open-source models can run locally without per-token usage fees, making them more accessible for small law firms, startups, and other budget-conscious organizations. 
\end{itemize}

\subsection{Implementation Details}
In this work, we formulate our primary task as identifying which substrings of a contract (i.e., clauses) correspond to specific label categories, each linked a type of legal risk. We prompt LLMs to act as junior legal assistants in a law firm to review contracts and identify clauses that are relevant the given question. Specifically, each contract is provided to the LLM in full as context, along with a set of targeted questions corresponding to each label category (e.g., “On what date is the contract effective?”). The model is asked to extract and return only the sentence(s) from the contract that directly address the question. If the model does not find relevant clauses, we instruct the model to respond with ``no related clause.” CUAD provides ground truth annotations for this task. For every contract and associated question, CUAD includes the exact substrings (if any) that should be highlighted as relevant. We use these annotations as the ground truth and compare them with the predictions generated by each LLM.
Figure~\ref{fig:workflow} illustrates the overall workflow of the contract review task.

\subsection{Evaluation Metrics}
To ensure that the evaluation align with practical legal tasks, we assess model performance from three perspectives: 

\noindent\textbf{Correctness of Risk Identification.} We evaluate the correctness of each LLM's predictions using standard classification metrics: F1 score, F2 score, given the notable imbalance between positive cases and negative cases in CUAD test data. These metrics are computed based on true positives (TP), false positives (FP), false negatives (FN), and true negatives (TN) defined as follows: 
    \begin{itemize}
        \item TP, the label is not empty, and the model’s prediction \emph{fully} covers the labeled span;
        \item TN, the label is empty, and the model correctly predicts ``no related clause'';
        \item FP, the label is empty, but the model incorrectly predicts a non-empty clause;
        \item FN, the label is not empty, but the model either outputs ``no related clause'' or fails to fully cover the label span.
    \end{itemize}

\noindent
The F1 score is defined as:
    \[
    F_1 = 2 \cdot \frac{\mathrm{Precision} \cdot \mathrm{Recall}}{\mathrm{Precision} + \mathrm{Recall}}
    \]
    The F2 score is defined as:
    \[
    F_2 = 5 \cdot \frac{\mathrm{Precision} \cdot \mathrm{Recall}}{4 \cdot \mathrm{Precision} + \mathrm{Recall}}
    \]
This evaluation perspective simulates how senior lawyers assess whether junior legal assistants correctly identify and flag all relevant clauses. 
    
    
    \noindent \textbf{Output Effectiveness and Conciseness.} We assess how effectively and concisely the model outputs match the label using the average Jaccard similarity coefficients on positive cases. The Jaccard similarity coefficient is defined as:
    \[
    J(A, B) = \frac{|A \cap B|}{|A \cup B|}
    \]
    where \( A \) and \( B \) are the sets of tokens in the predicted and ground truth spans, respectively \cite{Jaccard}. This metric reflects how senior lawyers evaluate the quality of juniors' summaries: accurate but not overly wordy.  Given the high hourly rates of senior lawyers (often ranging from \$2,000 to \$3,000) \cite{hourlyrate},  conciseness in identifying legal risks is crucial.
    
    \noindent \textbf{Laziness Detection.}  We measure how often LLMs avoid answering questions by directly outputting ``no related clause.'' Specifically, we calculate the false ``no related clause'' rate, defined as the proportion of test cases where the model outputs ``no related clause'' even though a non-empty clause exists in the ground truth. This is a subset of the broader false negative definition and focuses on identifying overly conservative or underperforming models that fail to retrieve relevant information.

\section{\ContractEval  Results Analysis}

In this section, we analyze the results from seven perspectives and address the following research questions (\textbf{RQs}):

\begin{itemize}
\item \textbf{RQ1:} How do proprietary models compare with open-source models in correctness of legal risk identification?
\item \textbf{RQ2:} To what extent do proprietary models outperform open-source models in output effectiveness?
\item \textbf{RQ3:} How often do LLMs exhibit ``laziness" by incorrectly returning "no related clause" responses?
\item \textbf{RQ4:} What is the impact of model size on performance in clause-level legal risk identification?
\item \textbf{RQ5:} How do reasoning strategies (``thinking" vs. ``non-thinking" mode) affect model performance?
\item \textbf{RQ6:} What are the effects of quantization on model performance and efficiency?
\item \textbf{RQ7:} How does performance vary across different categories of legal clauses?
\end{itemize}

\begin{table}[t]
\centering
\caption{Model Performance Comparison. Here, False rate refers to \emph{``No related Clause"} false rate. $\uparrow$ means the measure is the higher the better, while $\downarrow$ means the measure is the lower the better. Bold numbers indicate its performance is the best in proprietary models and open-source models.}
\label{Overalltable}
\begin{adjustbox}{width=\linewidth}
\setlength{\tabcolsep}{3pt}
\begin{tabular}{llccccc}
\toprule
\makecell{\textbf{Model Name}} & \makecell{\textbf{F1} $\uparrow$} & \makecell{\textbf{F2} $\uparrow$} & \makecell{\textbf{Jaccard} \\ \textbf{Similarity} $\uparrow$} & \makecell{\textbf{False} \\ \textbf{Rate} $\downarrow$} \\
\midrule
\multicolumn{5}{l}{\textbf{Proprietary Models}} \\
GPT 4.1 & 0.641 & 0.672 & 0.472 & 0.071 \\
GPT 4.1 mini & \textbf{0.644} & \textbf{0.678} & 0.435 & 0.072 \\
Gemini 2.5 Pro Preview & 0.497 & 0.604 & \textbf{0.506} & \textbf{0.011} \\
Claude Sonnet 4 & 0.523 & 0.578 & 0.458 & 0.025 \\
\midrule
\multicolumn{5}{l}{\textbf{Open-source Models}} \\
DeepSeek R1 Distill Qwen 7B &0.071 & 0.085 & 0.131 & 0.037  \\
DeepSeek R1 0528 Qwen3 8B & 0.475 & 0.464 & 0.404 & 0.100 \\
Llama 3 1.8B Instruct & 0.392 & 0.370 & 0.300 & 0.214 \\
Gemma 3 4B & 0.188 & 0.246 & 0.311 & \textbf{0.000}\\
Gemma 3 12B & 0.391 & 0.421 & \textbf{0.446} & 0.045 \\
Qwen3 4B & 0.411 & 0.362 & 0.337 & 0.211\\
Qwen3 4B (thinking) & 0.075 & 0.055 & 0.300 & 0.198 \\
Qwen3 8B AWQ & 0.475 & 0.393 & 0.303 & 0.306 \\
Qwen3 8B AWQ (thinking) & 0.187 & 0.150 & 0.374 & 0.125 \\
Qwen3 8B & 0.530 & 0.453 & 0.340 & 0.248\\
Qwen3 8B (thinking) & \textbf{0.540} & \textbf{0.512} & 0.391 & 0.110 \\
Qwen3 8B FP8 &0.491 & 0.411 & 0.313 & 0.285\\
Qwen3 8B FP8 (thinking) &0.307 & 0.263 & 0.399 & 0.105 \\
Qwen3 14B &0.473 & 0.418 & 0.400 & 0.174 \\
Qwen3 14B (thinking) &0.387 & 0.334 & 0.421 & 0.117  \\
\bottomrule
\end{tabular}
\end{adjustbox}
\end{table}

\subsection{Proprietary Models Outperform Open-source Models in Correctness} We demonstrate the experiment results in Table \ref{Overalltable}. Proprietary models consistently outperform open-source models in correctness (as measured by F1 and F2 scores). GPT 4.1 and GPT 4.1 mini, achieve the highest F1 scores (0.641 and 0.644, respectively) and F2 scores (0.672 and 0.678), clearly leading the benchmark. These results show their ability to accurately identify and extract relevant clauses. The other two proprietary models, Claude Sonnet 4 and Gemini 2.5 Pro, also outperform most open-source models. 

Among open-source models, Qwen3 8B variants perform best in correctness. In particular, Qwen3 8B in ``thinking” mode achieves an F1 score of 0.540 and an F2 score of 0.512, still about 16\% lower than GPT 4.1. Qwen3 8B without thinking mode and FP8 variant also demonstrate competitive performance. However, most other open-source models fall far behind. For instance, DeepSeek R1 Distill Qwen 7B and Qwen3 4B in ``thinking” mode score below 0.1 in F1, indicating near-total failure on this task. These results suggest that smaller or instruction-optimized open models, particularly those lacking domain-specific tuning, are currently ill-equipped for accurate clause-level legal risk identification.

\subsection{Proprietary Models Also Excel in Output Effectiveness and Accuracy}
Beyond correctness, proprietary models also demonstrate better output effectiveness measured by Jaccard similarity. This metric reflects how precisely models extract relevant clause spans without including extra content. As shown in Table \ref{Overalltable}, most proprietary models achieve higher Jaccard similarity scores compared to open-source models. Notably, GPT 4.1 and Claude Sonnet 4 both maintain Jaccard scores above 0.45, indicating strong overlap between predicted outputs and ground-truth annotations.

One exception is the open-source model Gemma 3 12B, which achieves a surprisingly competitive Jaccard similarity of 0.446, higher than GPT 4.1 mini and close to GPT 4.1. However, this result is not a common case; most other open-source models perform much worse in Jaccard similarity. For instance, although Qwen3 8B variants perform reasonably well in terms of F1 and F2 scores, their Jaccard similarity drops to 0.391 or lower. This suggests that while these models may correctly identify relevant clauses, they often include surrounding or irrelevant content.

Overall, the lower Jaccard scores among open-source models point to a tendency to over-select or include non-essential text when extracting clauses. This behavior negatively impacts output effectiveness, especially in the legal area where concise and accurate clause extraction is essential. Thus, minimizing irrelevant content extraction remains a key area for improving the practical deployment of open-source LLMs in legal document review.


\subsection{Missing Relevant Legal Risks} 
\begin{figure}[t]
    \centering
    \includegraphics[width=\linewidth]{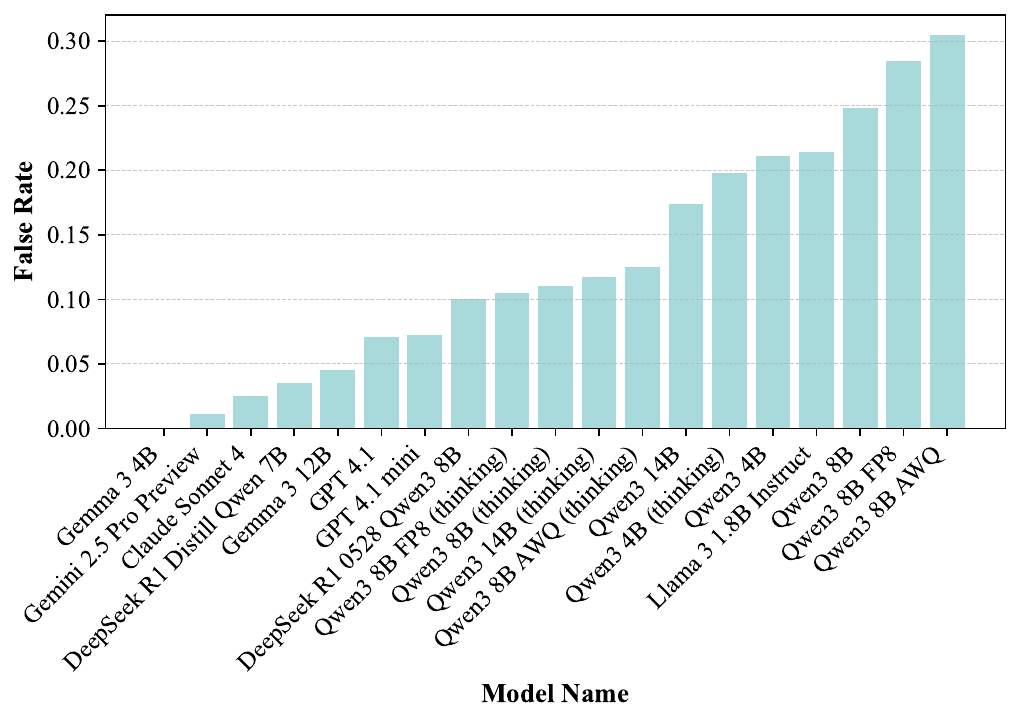}
    \caption{False ``No Related Clause" Rate by Model}
    \label{fig:fnr}
\end{figure}

In addition to correctness and output effectiveness, we examine how often models incorrectly return “no related clause” when a relevant clause is actually present. This metric serves as a proxy for a model's information retrieval capability, and high false “no related clause” rate may suggest model “laziness” or low confidence in clause extraction. As shown in Figure~\ref{fig:fnr}, performance varies widely across models. Notably, Gemma 3 4B achieves a false “no related clause” rate of 0.000, meaning it successfully retrieves relevant content in all applicable cases, an impressive result for a smaller open-source model. Proprietary models such as Gemini 2.5 Pro (0.011), Claude Sonnet 4 (0.025), and both GPT 4.1 variants (0.07) maintain low false rates.
In contrast, several open-source Qwen3 models, especially in non-thinking mode, exhibit substantially higher false rates. For instance, Qwen3 8B AWQ in non-thinking mode exceeds a 30\% false rate, indicating that nearly one-third of its answers miss relevant clauses entirely. Models like LLaMA 3.1 8B and Qwen3 8B FP8 in non-thinking mode also show concerningly high rates, ranging from 25–28\%. These results suggest a tendency among some models to skip applicable clauses, potentially undermining their practical value in legal review tasks.

Overall, minimizing false ``no related clause” responses is critical in contract review, because overlooking a clause may have serious consequences in legal services. Our findings highlight the importance of improving information retrieval capability in open-source models to ensure comprehensive contract risk analysis. 

\subsection{Impact of Model Size}  Among open-source models, scaling up generally led to better correctness, but the returns start to slow down or even diminish. As shown in Figure~\ref{fig:qwen3_comparison}, within the Qwen3 family, the 8B model achieves the highest F1 scores: 0.530 in non-thinking mode and 0.540 in thinking mode. Interestingly, this performance surpasses both the smaller Qwen3-4B and the larger Qwen3 14B model. This suggests that the 8B model reaches an optimal balance between capacity and task-specific generalization, at least under the current training and fine-tuning conditions.

However, the 14B model exhibits diminishing and even negative returns in F1 scores, despite its larger scale. This indicates larger models do not guarantee improved clause-level correctness, and may instead require more extensive domain-specific fine-tuning or adaptation to unlock its full potential. In contrast, in terms of output effectiveness, as measured by Jaccard similarity (Figure~\ref{fig:qwen3_comparison}), larger models generally perform better. This trend suggests that larger model may better capture relevant clause without adding too much extra content. Therefore, while correctness (F1 score) may level off, effectiveness (Jaccard similarity) continues to benefit from scaling.

\subsection{Reasoning Strategies: ``thinking" vs. ``non-thinking"} Some open-source models support two reasoning strategies, enabling and disabling the ``thinking" mode. Thinking mode is intended to encourage step-by-step reasoning. While such reasoning strategies are often beneficial for complex tasks, our results indicate mixed outcomes for clause-level legal risk identification.

\begin{figure}[t]
    \centering
    \includegraphics[width=0.99\linewidth]{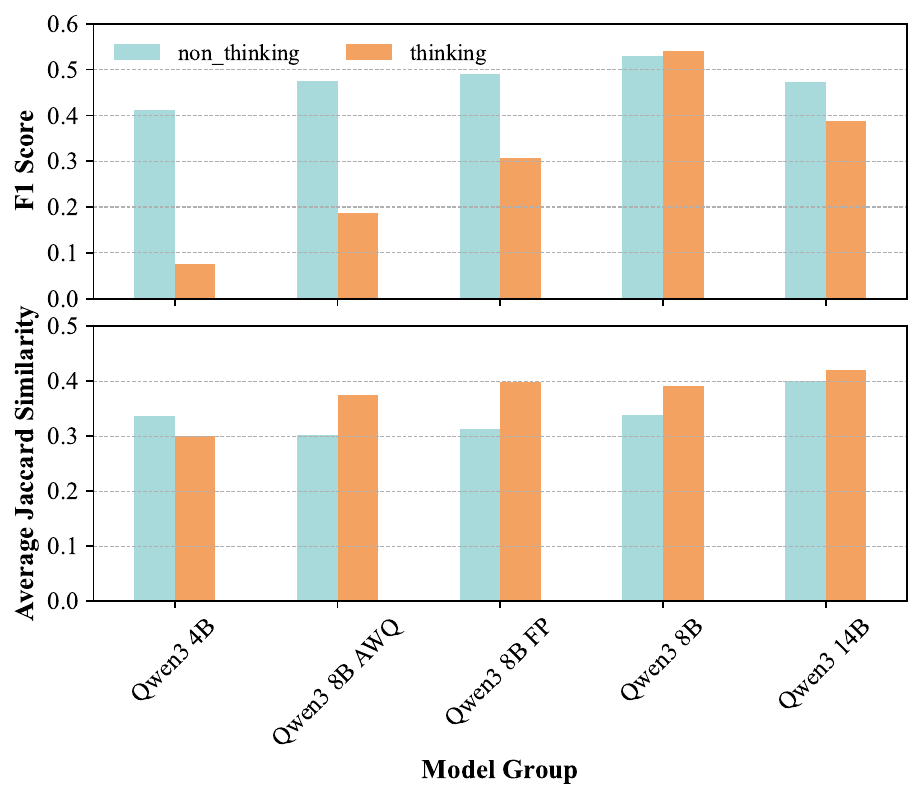}
    \caption{Qwen3 Model Series Performance}
    \label{fig:qwen3_comparison}
\end{figure}

As shown in Figures \ref{fig:qwen3_comparison}, thinking mode often improves output effectiveness, which is reflected in higher Jaccard similarity scores, but simultaneously reduces correctness, with notable drops in F1 scores across most models, including Qwen3 4B, 8B, and 14B. This trade-off suggests that reasoning-based prompting may lead models to over-explain or include irrelevant clauses, especially when dealing with relatively straightforward extraction tasks.

These findings point to a potential mismatch: while thinking mode encourages more deliberate generation, it may also increase verbosity or hallucination when the task only requires precise span identification. Additionally, it is likely that these reasoning-augmented models have not been pre-trained or fine-tuned on legal-specific tasks, limiting their effectiveness in high-precision extraction scenarios common in legal contract review.

\subsection{Performance Variation among Categories} 
\begin{figure*}[htbp]
    \centering
    \includegraphics[width=0.98\textwidth]{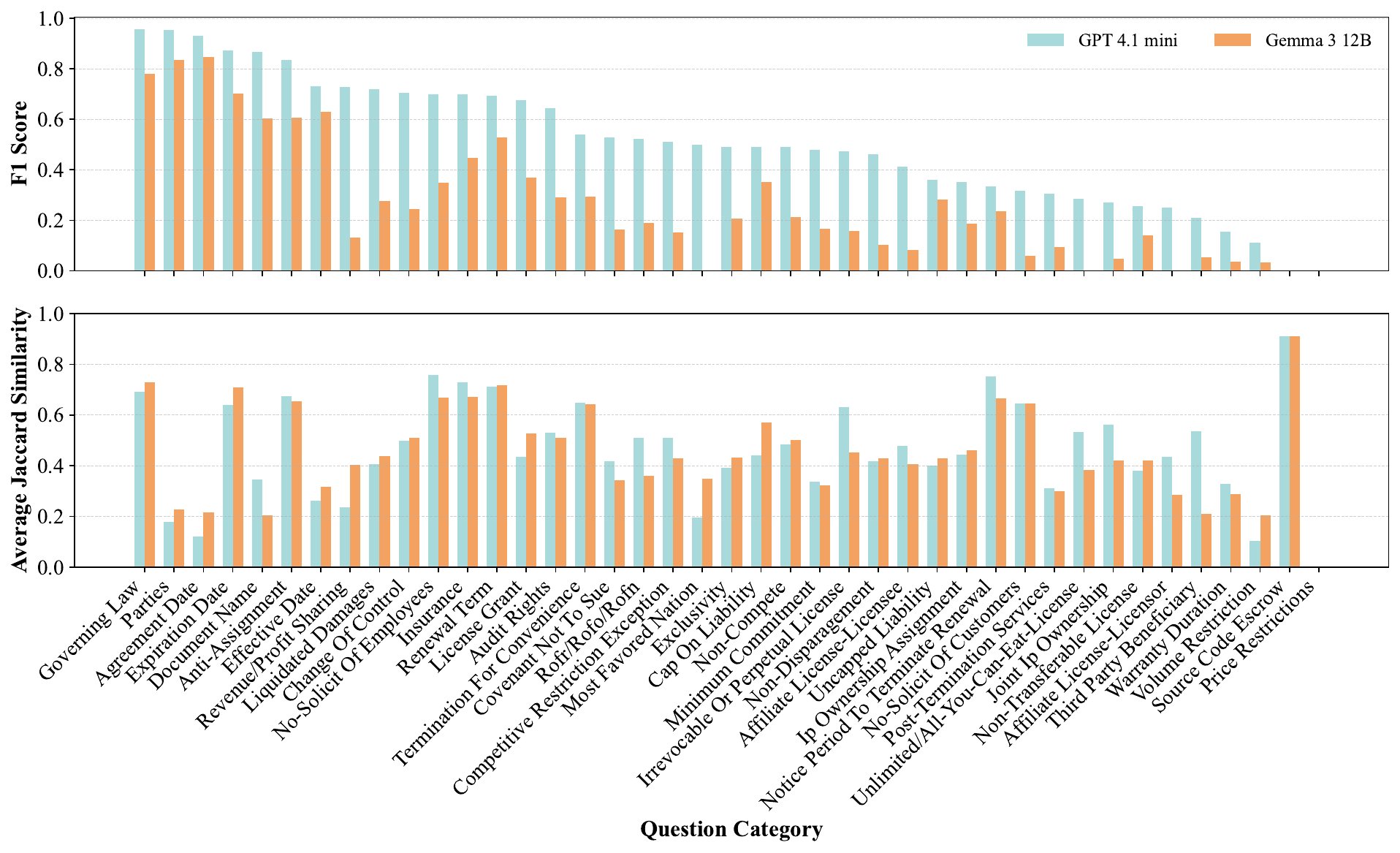}
    \caption{Performance by Question Category for GPT 4.1 mini and Gemma 3 12B}
    \label{fig:category_combined}
\end{figure*}

Figure \ref{fig:category_combined} compares F1 score and Jaccard similarity of GPT 4.1 mini with Gemma 3 12B across 41 clause categories (sorted based on the F1 score). We choose Gemma 3 12B because it is the open-source model with highest Jaccard similarity and also relatively low false ``no related clause". In the F1 score, GPT 4.1 mini consistently outperforms Gemma 3 12B across nearly all categories. In more straightforward tasks such as identifying ``Governing Law" and ``Parties", GPT 4.1 mini approaches or exceeds an F1 score of 0.9. However, both models show near-zero scores in more nuanced or rare categories like ``Uncapped Liability", ``Joint IP Ownership" and ``Notice Period to Terminate Renewal". But these clauses are important and of high risks in real life. This performance gap illustrates that while current LLMs excel in standard, frequently occurring clause types, they struggle with less common or more complex clauses.

The Jaccard similarity distribution reveals a more mixed performance. First, the Jaccard similarity score varies even within straightforward and simple clauses, as both models perform well on ``Governing Law" but fall short on ``Parties". Second, for more complex or less common clauses, Gemma 3 12B matches or slightly outperforms GPT 4.1 mini. For instance, Gemma 3 12B is better in ``IP Ownership” and``Source Code Escrow”. The narrower gap in Jaccard similarity indicates that top open-source models may retrieve partial clause content reasonably well and tend to include less irrelevant extra content.

Overall, the results underscore the need for fine-grained evaluation at the clause level. They also point to a critical area for improvement: improving model performance on less common but high-risk clause types. Addressing this imbalance is essential for practical deployment of LLMs in contract review, where missing a rare clause could have serious legal consequences.



\subsection{Effects of Quantization} Quantization is a useful efficiency tradeoff for deployment. For instance, the Qwen3 8B uses its default bfloat16 floating-point format, which takes double the GPU consumption as Qwen3 8B FP8 and four times the GPU consumption as Qwen3 8B AWQ. Comparing Qwen3 8B with the quantized version Qwen3 8B FP8, we find that quantization to FP8 results in a slight performance drop. However, when combined with thinking mode, quantized models exhibited more obvious degradation. These results indicate that quantization is a viable approach for inference efficiency in standard settings, but its harms to performance are more evident in reasoning models.

\section{Discussions: Takeaways for LLM-Aided Legal Practice and Future Research}
LLMs, particularly top proprietary models like GPT 4.1 and GPT 4.1 mini, are approaching the performance level of junior legal assistants: capable of identifying relevant clauses but still requiring oversight from senior professionals. While open-source models generally lag slightly behind, they demonstrate strong potential in specific dimensions such as output effectiveness and conciseness . Given the practical appeal of locally deployed open-source models, our findings suggest three key directions for improvement: (1) fine-tuning models to more accurately identify relevant clauses; (2) addressing category imbalance by focusing on rare or complex clauses; and (3) mitigating model ``laziness,'' where LLMs overuse the ``no related clause'' response, which risks overlooking important legal issues and increasing potential financial liability.

Our results also point to several open questions. First, how can evaluation methods for contract review and legal tasks be better aligned with the standards and expectations of senior legal professionals? Second, given the high prompt sensitivity of LLMs, how can we improve their reliability and consistency in real-world workflows? While these questions are beyond the scope of this benchmark, they are crucial for future progress and call for closer collaboration between the LLM research and legal communities.

\section{Conclusions}

In this paper, we present \ContractEval, the first benchmark for systematically evaluating proprietary and open-source LLMs on clause-level contract review. Using the CUAD dataset, we assess 19 state-of-the-art LLMs across dimensions such as correctness, output effectiveness, and false ``no related clause" rates. Our benchmarking results indicate that proprietary models still lead overall, but top open-source models are narrowing the gap in either output effectiveness or false ``no related clause" rates. We also observe that larger model size does not always guarantee better performance, and reasoning-based prompting can increase output conciseness at the cost of lowering correctness. These findings highlight the importance of targeted fine-tuning and robust benchmarking to advance open-source LLMs' performance on contract review and clause analysis.


\bibliographystyle{IEEEtran}
\bibliography{main}

\begin{appendices}

\onecolumn
\section{Full List of Label Categories and Questions}
\renewcommand{\arraystretch}{1.2}  
\begin{center}
\setlength{\LTcapwidth}{\textwidth}
\begin{longtable}{p{0.2\textwidth}|p{0.65\textwidth}}
\caption{Label Categories and Questions} \label{tab:cuad-longtable} \\
\hline
\textbf{Category} & \textbf{Detailed Question} \\
\hline
\endfirsthead

\hline
\textbf{Category} & \textbf{Detailed Question} \\
\hline
\endhead

\hline
\multicolumn{2}{r}{\textit{Continued on next page}} \\
\endfoot

\hline
\endlastfoot

Document Name & The name of the contract \\
\hline
Parties & The two or more parties who signed the contract \\
\hline
Agreement Date & The date of the contract \\
\hline
Effective Date & On what date is the contract effective? \\
\hline
Expiration Date & On what date will the contract’s initial term expire? \\
\hline
Renewal Term & What is the renewal term after the initial term expires? This includes automatic extensions and unilateral extensions with prior notice. \\
\hline
Notice to Terminate Renewal & What is the notice period required to terminate renewal? \\
\hline
Governing Law & Which state/country’s law governs the interpretation of the contract? \\
\hline
Most Favored Nation & Is there a clause that if a third party gets better terms on the licensing or sale of technology/goods/services described in the contract, the buyer of such technology/goods/services under the contract shall be entitled to those better terms? \\
\hline
Non-Compete & Is there a restriction on the ability of a party to compete with the counterparty or operate in a certain geography or business or technology sector? \\
\hline
Exclusivity & Is there an exclusive dealing commitment with the counterparty? This includes a commitment to procure all “requirements” from one party of certain technology, goods, or services or a prohibition on licensing or selling technology, goods or services to third parties, or a prohibition on collaborating or working with other parties), whether during the contract or after the contract ends (or both). \\
\hline
No-Solicit of Customers & Is a party restricted from contracting or soliciting customers or partners of the counterparty, whether during the contract or after the contract ends (or both)? \\
\hline
Competitive Restriction Exception & This category includes the exceptions or carveouts to Non-Compete, Exclusivity and No-Solicit of Customers above. \\
\hline
No-Solicit of Employees & Is there a restriction on a party’s soliciting or hiring employees and/or contractors from the counterparty, whether during the contract or after the contract ends (or both)? \\
\hline
Non-Disparagement & Is there a requirement on a party not to disparage the counterparty? \\
\hline
Termination for Convenience & Can a party terminate this contract without cause (solely by giving a notice and allowing a waiting period to expire)? \\
\hline
ROFR/ROFO/ROFN & Is there a clause granting one party a right of first refusal, right of first offer or right of first negotiation to purchase, license, market, or distribute equity interest, technology, assets, products or services? \\
\hline
Change of Control & Does one party have the right to terminate or is consent or notice required of the counterparty if such party undergoes a change of control, such as a merger, stock sale, transfer of all or substantially all of its assets or business, or assignment by operation of law? \\
\hline
Anti-Assignment & Is consent or notice required of a party if the contract is assigned to a third party? \\
\hline
Revenue/Profit Sharing & Is one party required to share revenue or profit with the counterparty for any technology, goods, or services? \\
\hline
Price Restriction & Is there a restriction on the ability of a party to raise or reduce prices of technology, goods, or services provided? \\
\hline
Minimum Commitment & Is there a minimum order size or minimum amount or units per time period that one party must buy from the counterparty under the contract? \\
\hline
Volume Restriction & Is there a fee increase or consent requirement, etc. if one party’s use of the product/services exceeds certain threshold? \\
\hline
IP Ownership Assignment & Does intellectual property created by one party become the property of the counterparty, either per the terms of the contract or upon the occurrence of certain events? \\
\hline
Joint IP Ownership & Is there any clause providing for joint or shared ownership of intellectual property between the parties to the contract? \\
\hline
License Grant & Does the contract contain a license granted by one party to its counterparty? \\
\hline
Non-Transferable License & Does the contract limit the ability of a party to transfer the license being granted to a third party? \\
\hline
Affiliate IP License Licensor & Does the contract contain a license grant by affiliates of the licensor or that includes intellectual property of affiliates of the licensor? \\
\hline
Affiliate IP License Licensee & Does the contract contain a license grant to a licensee (incl. sublicensor) and the affiliates of such licensee/sublicensor? \\
\hline
Unlimited/All-You-Can-Eat License & Is there a clause granting one party an “enterprise,” “all you can eat” or unlimited usage license? \\
\hline
Irrevocable or Perpetual License & Does the contract contain a license grant that is irrevocable or perpetual? \\
\hline
Source Code Escrow & Is one party required to deposit its source code into escrow with a third party, which can be released to the counterparty upon the occurrence of certain events (bankruptcy, insolvency, etc.)? \\
\hline
Post-Termination Services & Is a party subject to obligations after the termination or expiration of a contract, including any post-termination transition, payment, transfer of IP, wind-down, last-buy, or similar commitments? \\
\hline
Audit Rights & Does a party have the right to audit the books, records, or physical locations of the counterparty to ensure compliance with the contract? \\
\hline
Uncapped Liability & Is a party’s liability uncapped upon the breach of its obligation in the contract? This also includes uncap liability for a particular type of breach such as IP infringement or breach of confidentiality obligation. \\
\hline
Cap on Liability & Does the contract include a cap on liability upon the breach of a party’s obligation? This includes time limitation for the counterparty to bring claims or maximum amount for recovery. \\
\hline
Liquidated Damages & Does the contract contain a clause that would award either party liquidated damages for breach or a fee upon the termination of a contract (termination fee)? \\
\hline
Warranty Duration & What is the duration of any warranty against defects or errors in technology, products, or services provided under the contract? \\
\hline
Insurance & Is there a requirement for insurance that must be maintained by one party for the benefit of the counterparty? \\
\hline
Covenant Not to Sue & Is a party restricted from contesting the validity of the counterparty’s ownership of intellectual property or otherwise bringing a claim against the counterparty for matters unrelated to the contract? \\
\hline
Third Party Beneficiary & Is there a non-contracting party who is a beneficiary to some or all of the clauses in the contract and therefore can enforce its rights against a contracting party? \\
\hline
\end{longtable}
\end{center}

\newpage         
\twocolumn 

\section{Examples of Prompt}
As illustrated in Figure \ref{fig:workflow}, we ask LLMs to act as legal assistant helping with contract review task to extract relevant clauses in the context. The extraction is guided by the following system prompt:

\begin{quote}
\textit{
You are an assistant with strong legal knowledge, supporting senior lawyers by preparing reference materials.
Given a Context and a Question, extract and return only the sentence(s) from the Context that directly address or relate to the Question.
Do not rephrase or summarize in any way—respond with exact sentences from the Context relevant to the Question. If a relevant sentence contains unrelated elements such as page numbers or whitespace, include them exactly as they appear.
If no part of the Context is relevant to the Question, respond with: "No related clause.
}
\end{quote}

\end{appendices}


\end{document}